\documentclass[sigconf]{acmart}

\AtBeginDocument{%
  \providecommand\BibTeX{{%
    \normalfont B\kern-0.5em{\scshape i\kern-0.25em b}\kern-0.8em\TeX}}}

\setcopyright{acmcopyright}
\copyrightyear{2022}
\acmYear{2022}
\acmDOI{XXXXXXX.XXXXXXX}

\acmConference[MM' 22]{ACM Multimedia}{October 10--14, 2022}{Lisbon, Protugal}
\acmPrice{15.00}
\acmISBN{978-1-4503-XXXX-X/18/06}

\usepackage{multirow}

\newcommand{\figref}[1]{\mbox{Fig.~\ref{#1}}}
\newcommand{\tabref}[1]{\mbox{Table~\ref{#1}}}
\renewcommand{\eqref}[1]{\mbox{Eqn.~\ref{#1}}}

\newcommand{\etal}{\textit{et al}.}
\newcommand{\ie}{\textit{i}.\textit{e}.,}
\newcommand{\eg}{\textit{e}.\textit{g}.,}

\newcommand{\blue}[1]{\textcolor{blue}{#1}}

\begin{document}

\title{A Unified End-to-End Retriever-Reader Framework for Knowledge-based VQA}



\author{\mbox{
Yangyang Guo$^\dag$,  
Liqiang Nie$^{\star}$, 
Yongkang Wong$^\dag$, 
Yibing Liu$^\ddag$, 
Zhiyong Cheng$^{\S}$, 
Mohan Kankanhalli$^\dag$}}

\affiliation{
\institution{
$^\dag$National University of Singapore, $^\star$Harbin Institute of Technology (Shenzhen), \\ $^\ddag$City University of Hong Kong, $^\S$Qilu University of Technology (Shandong Academy of Sciences)}
\country{}
}

\email{{guoyang.eric, nieliqiang, lyibing112, jason.zy.cheng}@gmail.com, {yongkang.wong, mohan}@comp.nus.edu.sg}

\renewcommand{\shortauthors}{}
\settopmatter{printacmref=false}


\begin{abstract}

Knowledge-based Visual Question Answering (VQA) expects models to rely on external knowledge for robust answer prediction. Though significant it is, this paper discovers several leading factors impeding the advancement of current state-of-the-art methods. On the one hand, methods which exploit the explicit knowledge take the knowledge as a complement for the coarsely trained VQA model. Despite their effectiveness, these approaches often suffer from noise incorporation and error propagation. On the other hand, pertaining to the implicit knowledge, the multi-modal implicit knowledge for knowledge-based VQA still remains largely unexplored. This work presents a unified end-to-end retriever-reader framework towards knowledge-based VQA. In particular, we shed light on the multi-modal implicit knowledge from vision-language pre-training models to mine its potential in knowledge reasoning. As for the noise problem encountered by the retrieval operation on explicit knowledge, we design a novel scheme to create pseudo labels for effective knowledge supervision. This scheme is able to not only provide guidance for knowledge retrieval, but also drop these instances potentially error-prone towards question answering. To validate the effectiveness of the proposed method, we conduct extensive experiments on the benchmark dataset. The experimental results reveal that our method outperforms existing baselines by a noticeable margin. Beyond the reported numbers, this paper further spawns several insights on knowledge utilization for future research with some empirical findings.

\begin{figure}[!t]
  \centering
  \includegraphics[width=1.0\linewidth]{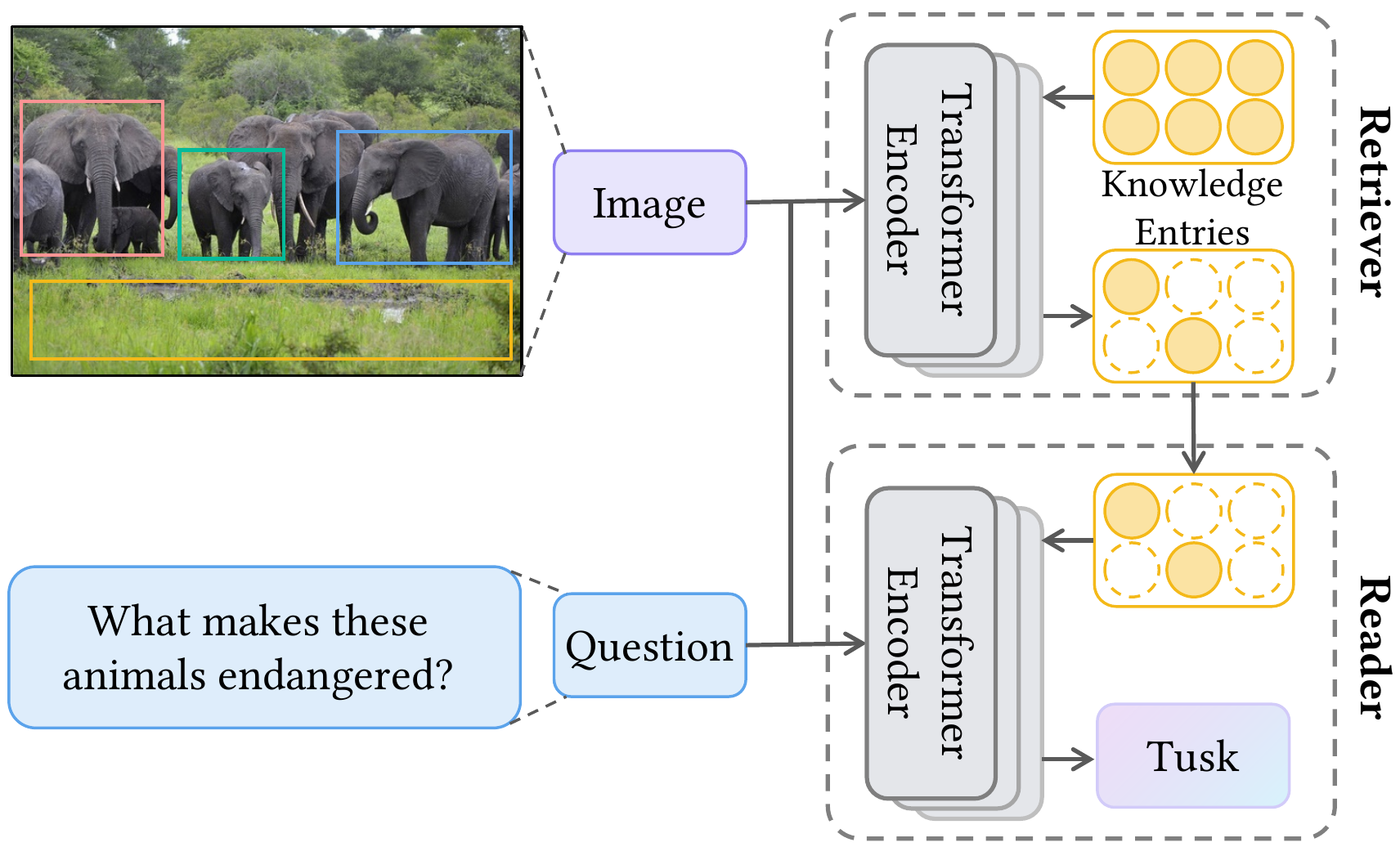}
  \vspace{-2em}
  \caption{An illustrative sketch of the proposed UnifER. The Retriever is leveraged to filter the large explicit knowledge entry set to a relevant subset, which is essential for question answering. The retrieved subset, together with the question and image, is then taken as inputs for answer prediction by the Reader. Note that the multi-modal implicit knowledge is freely embodied in the bottom Transformer Encoder.}\label{fig:teaser}
  \vspace{-2em}
\end{figure}

\end{abstract}



\maketitle

\vspace{-1em}
\subsection*{\textbf{ACM Reference Format:}}{Yangyang Guo, Liqiang Nie, Yongkang Wong, Yibing Liu, Zhiyong Cheng and Mohan Kankanhalli. 2022. A Unified End-to-End Retriever-Reader Framework for Knowledge-based VQA. In Proceedings of ACM Multimedia (MM’ 22). ACM, New York, NY, USA, 10 pages. http}

\section{Introduction}\label{sec:introduction}

Visual Question Answering (VQA) offers to perform visual reasoning at the junction of vision and language. Despite its remarkable progress over the years, existing VQA benchmarks mostly address simple recognition questions (\eg~\emph{what color}) while struggling on the ones requiring common sense and domain knowledge beyond the given image\footnote{One exception is the GQA dataset~\cite{gqa}, which offers to build compositional and multi-hop reasoning questions.}. To settle this concern, the task of knowledge-based VQA~\cite{ok-vqa}, serving as a proxy to understand the open-ended scenes, has recently been advocated to bridge this gap. The challenging nature of the knowledge-hungry questions makes conventional VQA models less reliable~\cite{ok-vqa, mm}. For example, the question in \figref{fig:teaser} asks the reason of why elephants have become endangered. 
Answering this question is rather tricky, only if one retrieves the pertinent knowledge -- \emph{poachers hunting elements for tusk}, where most prior art fails. 

In general, knowledge studied in literature can be roughly categorized as explicit and implicit. Existing studies often take explicit knowledge as a complementary clue for answer prediction~\cite{conceptbert, mm, ok-vqa}, such as passages in Wikipedia~\cite{wiki} and $<$subject, relation, object$>$ triplets in ConceptNet~\cite{conceptnet}. Regarding the training procedure, in addition to the question-image branch, a knowledge branch is introduced and combined with the former  by a late fusion manner~\cite{ok-vqa, conceptbert}. Nevertheless, this strategy is hampered by two downsides: 1) Injecting more knowledge to VQA models will unavoidably lead to more noises as well as an increased retrieval overhead, and 2) since there is no supervision for knowledge retrieval, the estimated empirical errors will be propagated when the retrieved knowledge becomes inapplicable (\eg~\emph{tree is green} for the question in \figref{fig:teaser}).

Distinct from the above methods, recent approaches have shed light on the extensive implicit knowledge from large-scale pre-trained language models, such BERT~\cite{bert} and GPT-3~\cite{gpt-3}. To this end, they have devoted many efforts to transferring the knowledge to guide the common sense reasoning in VQA models. 
Although achieving good results, these methods only make attempts to the pre-trained language model. With the prevalence of vision-language pre-training, whether multi-modal implicit knowledge from this research benefits knowledge-based VQA remains an open question. 

In this work, we make contributions towards solving the aforementioned problems from three aspects. Overall, a \textbf{Unif}ied \textbf{E}nd-to-end \textbf{R}etriever-reader framework, dubbed as \textbf{UnifER}, is presented for end-to-end training in knowledge-based VQA. As shown in \figref{fig:teaser}, both image-question and knowledge entries (knowledge triplets can be converted into sentences, e.g., <elephant, related to, tusk> to ‘elephant is related to tusk’) are embedded by a Transformer Encoder~\cite{transformer}, which are then matched by a pre-defined similarity function in the Retriever. After retrieving the most relevant knowledge entries, UnifER performs answer prediction with the Reader. Secondly, our Transformer Encoder (especially the bottom one) is inspired by the recent advancement in vision-language pre-training~\cite{visual-bert, lxmert, vilt}. To smoothly benefit from the multi-modal implicit knowledge, we employ these pre-trained models to initialize parameters for multi-modal fusion. The final aspect from our method addresses the noise and error propagation problems raised by the inapplicable explicit knowledge. To this end, we design a novel scheme to simultaneously achieve the following two goals: relevance labeling for Retriever, and instance selection for Reader. Specifically, we adopt the mathematical difference between the answer prediction loss from $<$image, question, knowledge$>$, \ie~$\mathcal{L}_{QIK}$, and $<$image, question$>$, \ie~$\mathcal{L}_{QI}$, as a surrogate. When {\small $\mathcal{L}_{QI}-\mathcal{L}_{QIK} > 0$}, it is practical to assume that the retrieved knowledge at least yields a positive effect for predicting answers. Accordingly, the value produces a marked effect for both Retriever and Reader, since a negative value often corresponds to inapplicable knowledge regarding the given question.  

To validate the effectiveness of the proposed method, we perform extensive experiments on the open-domain knowledge-based VQA benchmark -- OK-VQA~\cite{ok-vqa}. As observed from the experimental results, our method surpasses the state-of-the-art baselines by a significant margin. In addition, our empirical findings pinpoint two spots that are somehow ignored by previous studies: 1) Implicit knowledge from vision-language pre-training models manifests more positive than the explicit one. 2) Knowledge retrieval demands supervision for improving answer prediction. 

In summary, the key contributions of this paper are three-fold:
\begin{itemize}
    \item We present a unified end-to-end retriever-reader framework towards knowledge-based VQA. Unlike prior methods, ours is able to be trained in an end-to-end fashion.
    \item For explicit knowledge incorporation, we design a novel scheme to overcome  noise and error propagation problems. We also explore the potential of multi-modal implicit knowledge for knowledge-based VQA, which is distinctive from existing methods based on pre-trained language models. 
    \item We conduct detailed experiments on the benchmark dataset and find that our method can effectively outperform the baselines. We release our codes to facilitate the reproducibility of this work\footnote{https://github.com/guoyang9/UnifER.}. 
\end{itemize}
\section{Related Work}\label{sec:related_work}

\subsection{Knowledge in VQA}
Existing studies on  knowledge-based VQA are relatively few due to its challenging nature. With the introduction of the Fact-based VQA (FVQA) dataset~\cite{fvqa}, some studies attempt to reason about an image on the basis of supporting facts~\cite{fvqa, predict_relation}. As the ground-truth knowledge triplets are annotated, earlier work focuses more on the retrieval operation, such as encoding the image and question with a knowledge base query~\cite{fvqa} or neural networks~\cite{predict_relation}. Subsequent research recognizes that the $<$subject, relation, object$>$ triplets naturally conform to the graph structure, \ie~subjects and relations to graph nodes while relations to graph edges. In view of this, the effectiveness of graph convolutional neural networks~\cite{gcn-rec} has been studied in knowledge-based VQA~\cite{gcn_fvqa, gat-scene-graph} due to their advances in other domains. Recently, Marino~\etal~\cite{ok-vqa} presented the OK-VQA dataset, which is the first of its kind in the form of questions requiring open-ended external knowledge. Different from FVQA~\cite{fvqa}, OK-VQA does not provide any closed set knowledge source or relevant supporting facts. To this end, current approaches mainly resort to explicit structured knowledge like ConceptNet~\cite{conceptnet}, unstructured knowledge WikiPedia~\cite{wiki}, or both, for enriching the knowledge reasoning capability~\cite{conceptbert, mm}. Implicit knowledge from pre-trained language models has also been exploited~\cite{krisp, mavex}.

Although the existing methods have made some progress on the benchmark dataset, the effectiveness of  multi-modal implicit knowledge from vision-language pre-training models still remains less explored. In this work, we proposed an end-to-end framework to employ this type of knowledge and also address the error propagation problem hindered by previous methods. 

\subsection{Vision-Language Pre-training}
The success of BERT pre-training~\cite{bert} in the language domain paves the way for the prosperity of Vision-Language Pre-training (VLP) over the past few years. Based on the model structures, current image text pre-training methods can be split into two groups: two-stream and single-stream. Studies from the former group employ the Transformer encoder~\cite{transformer} to encode the question and image separately~\cite{lxmert, vilbert}. The features from these two branches are then fused in following stages. In contrast, single-stream methods take image and question as inputs in a single step~\cite{visual-bert, vlbert, vilt}. Accordingly, the visual and textual features are interacted and fused in advance. One essential factor to train a VLP model is the design of pretext tasks, where typical ones include masked language modeling~\cite{bert}, masked image modeling~\cite{visual-bert, vlbert}, and image text matching~\cite{vlbert, vilt}. The former two intend to rebuild the masked elements based on the collaborative signals from the other modality while the last one aims to model the coherence between the two modalities.

\begin{figure*}[!t]
  \centering
  \includegraphics[width=0.95\linewidth]{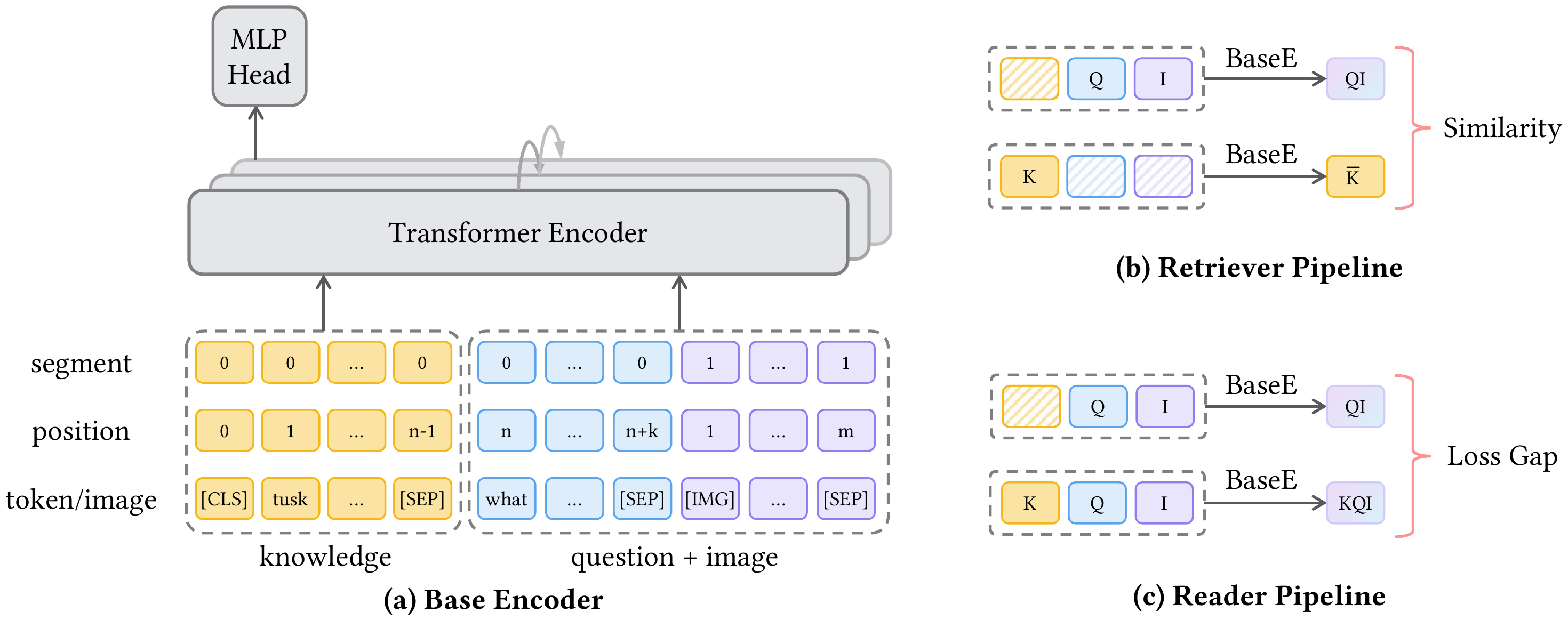}
  \vspace{-1em}
  \caption{Illustration of the Base Encoder and pipelines from the Retriever and Reader. (a) Base Encoder structure overview. The textual knowledge, together with the given question and image constitutes the inputs to our UnifER. After encoded with three kinds of embeddings, we leverage the Transformer Encoder to perform effective modal fusion. (b) Retriever Pipeline. The $<$question, image$>$ and knowledge are separately encoded by the Base Encoder. Cosine similarity is then adopted as signals for retrieving the most relevant knowledge entries. (c) Reader Pipeline. We expect that the loss from $QI$ is larger than that of $KQI$, in other words, the retrieved knowledge should at least make a positive effect to question answering.}\label{fig:structure}
  \vspace{-1em}
\end{figure*}

\subsection{Open Domain Question Answering}
Open Domain Question Answering (ODQA) has received considerable attention recently. Similar to the knowledge-based VQA, the objective of ODQA is to produce an answer for a question without accompanying documents that contain the right answer. To achieve this goal, an ODQA model usually consists of two components~\cite{end-odqa}: a retriever -- retrieving useful text pieces from open-ended knowledge sources; and a reader -- answering question given the filtered documents. A popular option for retriever is to embed the question and knowledge with the same encoder~\cite{bert, roberta}, where the similarity between these two can be estimated by a function like cosine similarity~\cite{retriever-odqa}. Subsequently, a machine reading comprehension model is often appended to generate the correct answer conditioned on the retrieved documents~\cite{mrc}. With the increasing computational resources, recent efforts contribute to train the retriever and reader in an end-to-end way, \eg~iteratively training retriever and reader~\cite{dpr1, dpr2} or by the expectation-maximization mechanism~\cite{end-odqa}.
\section{Method}\label{sec:method}

\subsection{Problem Formulation}
We consider a general setting of Visual Question Answering (VQA) under the open-ended scenes. Given a textual question $Q$ regarding an image $I$, the objective for knowledge-based VQA is given by,
\begin{equation} \label{equ:objective}
    \hat{a} = \arg \max_{a \in \Omega} p(a|Q, I; \mathbf{\Theta}),
\end{equation} 
where $\Omega$ and $\mathbf{\Theta}$ denote the answer set and model parameters, respectively. It is worth noting that the implicit knowledge can be inherently expressed by $\mathbf{\Theta}$, which is often extensively pre-trained on other large-scale datasets~\cite{concept-caption} with some dedicated pretext tasks~\cite{lxmert, vilt}. When it refers to the explicit knowledge, \eqref{equ:objective} can be decomposed as,
\begin{equation}
    p(a|Q, I, \mathcal{K}; \mathbf{\Theta}) =  \underbrace{p(\mathcal{K}_{ret}|Q, I, \mathcal{K}; \mathbf{\Phi})}_{\text{Retriever}} \cdot \underbrace{p(a|Q, I, \mathcal{K}_{ret}; \mathbf{\Theta})}_{\text{Reader}},
\end{equation}
where $\mathcal{K}$ and $\mathcal{K}_{ret}$ represent the overall broad knowledge set and the retrieved relevant subset, respectively. $\mathbf{\Phi}$ denotes the parameters from the model in the retrieval step. In particular, we refer the two components in knowledge-based VQA as Retriever and Reader. The former is to retrieve knowledge from $\mathcal{K}$, and the latter acts upon answer prediction with the retrieved knowledge $\mathcal{K}_{ret}$. 

Previous methods often adopt some pre-defined extraction approaches to obtain $\mathcal{K}_{rel}$, and emphasize more on the question answering target. In this work, we develop a novel end-to-end training framework to simultaneously perform retrieving and answering in a single step, which is elaborated in following subsections.

\subsection{Base Encoder for Knowledge-based VQA}
Before delving into the details of the proposed UnifER, we first present a Base Encoder structure, which constitutes an essential part for both Retriever and Reader in UnifER. The Base Encoder is enlightened by the recent prevalent VLP models~\cite{visual-bert, lxmert, vilt}, which demonstrate striking performance improvement in various vision-language tasks. In general, the overview of the proposed Base Encoder is shown in \figref{fig:structure}(a). 

\textbf{Input Processing.} There are three types of inputs to the Base Encoder: knowledge entry, question and image. For each structured knowledge entry such as $<$tusk, related to, weapon$>$, we transform the triplet to a textual sentence \emph{tusk related to weapon} such that it can be smoothly processed by the Transformer Encoder. Besides, each image can be divided into $32 \times 32$ patches, \ie~$\mathbf{V} \in \mathbb{R}^{32 \times 32 \times 3}$ following the ViT model~\cite{vit}; or we can employ the regional bottom-up features~\cite{updn} as raw input, \ie~$\mathbf{V} \in \mathbb{R}^{m \times b}$, where $m$ and $b$ imply the dimension of extracted visual features and the number of detected objects, respectively. 

\textbf{Modal Embedding.} Based on the post-processed knowledge entry, we first add two special tokens to the knowledge sentence $T_K$,
\begin{equation*}
    T = [CLS] \oplus T_K \oplus [SEP],
\end{equation*}
where $\oplus$ represents the concatenation operation; $[CLS]$ and $[SEP]$ are the classification token and separation token originally designed by BERT~\cite{bert}, respectively. We then append the textual question $T_Q$ to $T$ as,
\begin{equation*}
    T = T \oplus T_Q \oplus [SEP].
\end{equation*} 
Thereafter, we input the text $T$ to a pre-trained lookup table~\cite{bert, roberta} and obtain its word embedding $\mathbf{E}_w \in \mathbb{R}^{(n+k) \times d}$, where $n+k$ is the total number of tokens and $d$ represents the embedded dimension. Meanwhile, for the image input $\mathbf{V}$, we transform it to the same latent space with text $T$ as $\mathbf{E}_v \in \mathbb{R}^{m \times d}$ and concatenate it with $\mathbf{E}_w$ to obtain $\mathbf{E}_e \in \mathbb{R}^{(n+k+m) \times d}$.

In addition to the token/image embedding, we can also have a position embedding  $\mathbf{E}_p \in \mathbb{R}^{(n+k+m) \times d}$, which captures the sequential properties of inputs. Note that the position tokens for image can be static as there is no sequential patterns for an image. Besides, a segment embedding $\mathbf{E}_s \in \mathbb{R}^{(n+k+m) \times d}$ is introduced to indicate it is an image embedding as opposed to a textual embedding. Finally, we combine these three embeddings to a single one,
\begin{equation}
     \mathbf{E} =  \mathbf{E}_e +  \mathbf{E}_p +  \mathbf{E}_s.
\end{equation}

\textbf{Modal Fusion.} The past few years have witnessed the success of Transformer in multi-modal fusion~\cite{vilbert, lxmert, vilt}, which is distinguished from the conventional addition or concatenation techniques. Broadly speaking, these methods can be roughly categorized into two-stream and single-stream groups. Two-stream ones leverage a separate Transformer to perform intra-modal fusion for each modality, and then fuse the multi-modal inputs with another Transformer. By contrast, single-stream methods present to fuse different modality data in an early-fusion strategy. In other words, there is no discrimination between modalities after embedding. In what follows, we take the single-stream ones as an example to illustrate the modal fusion mechanism. 

Our Base Encoder consists of several stacked blocks, wherein each block includes a multi-head self-attention (MSA) layer and a multi-layer perceptron (MLP) layer. Besides, each layer also involves a layer normalization (LN) operation and residual connection. The $l$-th block of the encoder is given as,
\begin{equation}
    \left\{ 
    \begin{aligned}
    \mathbf{E}^0 &= \mathbf{E}, \\
    \hat{\mathbf{E}}^l &= \text{MSA}(\text{LN}(\mathbf{E}^{l-1})) + \mathbf{E}^{l-1}, \\
    \mathbf{E}^{l} &= \text{MLP}(\text{LN}(\hat{\mathbf{E}}^{l})) + \hat{\mathbf{E}}^l.
    \end{aligned}
    \right.
\end{equation}
Thereafter, we adopt the output from the last block $L$ for further estimation,
\begin{equation}
    \mathbf{o} = \text{MLP}(\mathbf{E}^{L}_{CLS}),
\end{equation}
where $\mathbf{E}^{L}_{CLS}$ denotes the output from the position of the $[CLS]$ token. In a nutshell, we employ a single function to represent the above procedures, 
\begin{equation}
    \mathbf{o} = f^{be}(Q, I, K; \mathbf{\Psi}),
\end{equation}
where $\mathbf{\Psi}$ implies all the involved parameters. 

There are three properties regarding our Base Encoder worth noting: 1) The three inputs are not necessarily required for one model. For instance, it can involve only the knowledge input, where the \emph{modal fusion} is accordingly reduced to the context of intra-modal. 2) Parameters in the Base Encoder with multi-modal inputs are initialized with vision-language pre-trained models, \eg~ViLT~\cite{vilt}. In this way, the multi-modal implicit knowledge can be seamlessly incorporated. For text-only input, like the knowledge entry, we employ the pre-trained Roberta~\cite{roberta} model for the Base Encoder. And 3) the Base Encoder serves as a building block for both the Retriever and Reader. Next, we will elaborate how we implement UnifER with it in more detail. 

\begin{figure}[!t]
  \centering
  \includegraphics[width=0.9\linewidth]{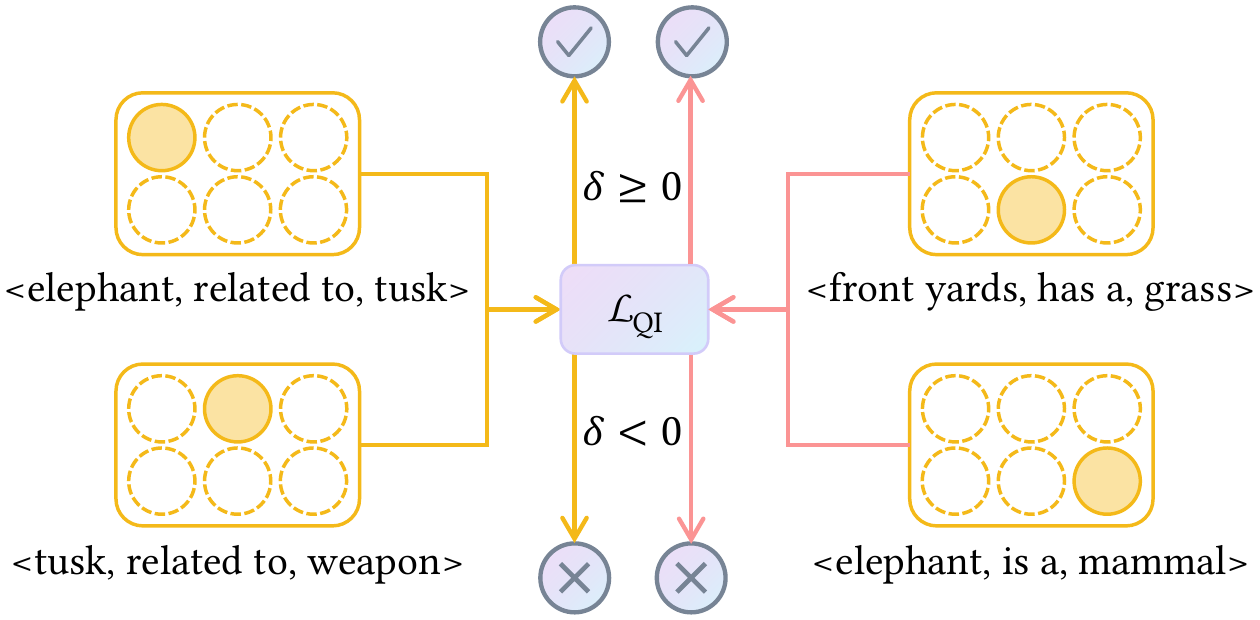}
  \vspace{-1em}
  \caption{Instance selection condition for Reader. According to the example in \figref{fig:teaser}, the left two knowledge entries are taken as applicable for the given image-question pair while the right two are not. When the mathematical difference value $\delta$ in \eqref{equ:valid} becomes negative, the instance will be ignored, otherwise it will be triggered for optimization.} \label{fig:condition}
  \vspace{-2em}
\end{figure}

\begin{table*}[!t]
    \centering
    \caption{Overall performance comparison between our method and state-of-the-art knowledge-based VQA models on the OK-VQA dataset, pertaining to the knowledge sources: WP - WikiPedia, CN - ConceptNet, DP - DBPedia, VG - Visual Genome, and GI - Google Image. The multi-modal implicit knowledge is marked in blue. The results from the eleven knowledge sub-categories are also reported. All results are in \% and the best performance over the current column is highlighted in bold.}
    \vspace{-1em}
    \scalebox{0.98}{
    \begin{tabular}{l|l|c|ccccccccccc}
    \toprule
        Method                              & Knowledge         & Overall   & VT    & BCP   & OMC   &SR     & CF    & GHLC  & PEL   & PA    & ST    & WC    & Other     \\
    \midrule
    \midrule
        Q Only~\cite{ok-vqa}                & -                 & 14.93     & 14.63 & 14.19 & 11.78 & 15.94 & 16.92 & 11.91 & 14.02 & 14.28 & 19.76 & 25.74 & 13.51     \\
        MLP~\cite{ok-vqa}                   & -                 & 20.67     & 21.33 & 15.81 & 17.76 & 24.69 & 21.81 & 11.91 & 17.15 & 21.33 & 19.29 & 29.92 & 19.81     \\
        BAN~\cite{ban}                      & -                 & 25.17     & 23.79 & 17.67 & 22.43 & 30.58 & 27.90 & 25.96 & 20.33 & 25.60 & 20.95 & 40.16 & 22.46     \\
        MUTAN~\cite{mutan}                  & -                 & 26.41     & 25.36 & 18.95 & 24.02 & 33.23 & 27.73 & 17.59 & 20.09 & 30.44 & 20.48 & 39.38 & 22.46     \\
        ArticleNet~\cite{ok-vqa}            & WP                & 5.28      & 4.48  & 0.93  & 5.09  & 5.11  & 5.69  & 6.24  & 3.13  & 6.95  & 5.00  & 9.92  & 5.33      \\
        \hspace{2mm}+BAN~\cite{ok-vqa}      & WP                & 25.61     & 24.45 & 19.88 & 21.59 & 30.79 & 29.12 & 20.57 & 21.54 & 26.42 & 27.14 & 38.29 & 22.16     \\
        \hspace{2mm}+MUTAN~\cite{ok-vqa}    & WP                & 27.84     & 25.56 & 23.95 & 26.87 & 33.44 & 29.94 & 20.71 & 25.05 & 29.70 & 24.76 & 39.84 & 23.62     \\
        GAT~\cite{gat-scene-graph}          & CN                & 29.03     & 26.11 & 24.15 & 26.36 & 36.94 & 30.92 & 25.15 & 24.83 & 29.58 & 29.38 & 39.64 & 26.34     \\
        ConceptBERT~\cite{conceptbert}      & CN                & 33.66     & 30.38 & 28.02 & 30.65 & 37.85 & 35.08 & 32.91 & 28.55 & 35.88 & 32.38 & 47.13 & 31.47     \\
        KG-Aug~\cite{mm}                    & CN+WP             & 16.11     & 15.54 & 12.11 & 13.94 & 19.60 & 16.93 & 14.61 & 13.37 & 16.55 & 11.48 & 27.92 & 14.98     \\
        \hspace{2mm}+BAN~\cite{mm}          & CN+WP             & 26.71     & 24.65 & 21.59 & 22.42 & 34.75 & 28.67 & 23.97 & 21.97 & 27.75 & 23.28 & 38.85 & 24.29     \\
        KRISP~\cite{krisp}                  & CN+DP+VG          & 38.35     & -     & -     & -     & -     & -     & -     & -     & -     & -     & -     & -         \\
        MAVEx~\cite{mavex}                  & CN+WP+GI          & 39.20     & -     & -     & -     & -     & -     & -     & -     & -     & -     & -     & -         \\
    \midrule
        \multirow{3}{*}{UnifER (Ours)}      & CN+\blue{VB}      & 30.16     & 27.84 & 33.16 & 25.97 & 43.41 & 29.99 & 34.42 & 27.46 & 30.33 & 24.41 & 37.84 & 29.98     \\ 
                                            & CN+\blue{LXMERT}  & 40.94     & 32.36 & 34.36 & 32.14 & 45.20 & 35.71 & 35.67 & 33.41 & 35.53 & 28.21 & 44.73 & 33.18     \\
                                            & CN+\blue{ViLT}    & \textbf{42.13}    & \textbf{36.88}    & \textbf{36.98}    & \textbf{37.52}
                                                                & \textbf{47.58}    & \textbf{46.88}    & \textbf{40.85}    & \textbf{37.29}
                                                                & \textbf{43.40}    & \textbf{35.95}    & \textbf{54.57}    & \textbf{39.58}                             \\
    \bottomrule  
    \end{tabular}}
    \label{tab:overall}
    \vspace{-1em}
\end{table*}

\subsection{Retriever}
The key element in knowledge-based VQA models is to effectively retrieve the most relevant knowledge for the given image-question pair. The evidence in recent advanced methods from open-domain question answering has demonstrated that dense retrieval technique can surpass conventional BM-25 or TF-IDF~\cite{dpr1, end-odqa} by a large margin. Inspired by this, we also leverage the deep neural networks to perform feature extraction for retrieval with multi-modal data. In particular, each knowledge entry and image-question pair are separately encoded with the Based Encoder,
\begin{equation}
    \left\{ 
    \begin{aligned}
    \mathbf{h}_{QI} &= f^{be}(Q, I; \mathbf{\Phi}_{QI}), \\
    \mathbf{h}_{K_i}  &= f^{be}(K_i; \mathbf{\Phi}_{K}),
    \end{aligned}
    \right.
\end{equation}
where $\mathbf{\Phi}_{QI}$ and $\mathbf{\Phi}_{K}$ represent the parameters from image-question branch and knowledge branch\footnote{The two sets of parameters can be fully/partially shared.}, respectively. We then employ  cosine similarity to estimate the proximity between each knowledge entry $K_i$ and the given image-question pair $QI$,
\begin{equation}\label{equ:sim}
    \text{Sim}(QI, K_i) = \frac{\mathbf{h}_{QI}^T \mathbf{h}_{K_i}}{\lVert \mathbf{h}_{QI} \rVert_2 \cdot \lVert \mathbf{h}_{K_i} \rVert_2},
\end{equation}
where $\text{Sim}(QI, K_i)$ naturally lies in the range of $[-1, 1]$. We illustrate the Retriever pipeline in \figref{fig:structure}(b), where $K$ and $QI$ inputs are separately masked. It can be inferred that with \eqref{equ:sim}, we can retrieve the most plausibly relevant subset $\mathcal{K}_{ret} = \{K_1, K_2, ..K_t\}$ from the whole knowledge set $\mathcal{K}$. Note that  there is no supervision for knowledge retrieval, arguably, $\mathcal{K}_{ret}$ can introduce noise during model training. For instance, the two entries in \figref{fig:condition}, \ie~\emph{$<$front yards, has a, grass$>$ and $<$elephant, is a, mammal$>$} are inapplicable for the question in \figref{fig:teaser}. 

\subsection{Reader}
After retrieving the desired knowledge, a common way for answer prediction is to input the knowledge, question and image to Base Encoder for modal fusion. To achieve this goal, we map the fused multi-modal features to the answer space given the fact that current VQA models all view VQA as a classification problem, \ie
\begin{equation}
    \hat{\mathbf{y}} = \text{MLP}(f^{be}(Q, I, K_i; \mathbf{\Theta})),
\end{equation}
where the length of $\hat{\mathbf{y}}$ equals to the size of the answer set $\Omega$. We then employ the cross entropy loss for model parameter optimization,
\begin{equation} \label{equ:lqik}
    \mathcal{L}_{QIK} (Q, I, K_i; \mathbf{\Theta}) = \sum_{j=1}^{\mid \Omega \mid} -y_j \log \hat{y}_j,
\end{equation}
where $y_j$ denotes the ground-truth label. 

Most recent methods optimize a loss function similar to \eqref{equ:lqik} and have devoted great efforts to incorporating abundant and diverse knowledge sources. However, we argue that this operation raises two problems: large retrieval overhead and error propagation. Regarding the former, introducing more knowledge is sub-optimal, as the knowledge source is unlimited and more knowledge will unavoidably result in increased retrieval burden. For the latter, if the retrieved knowledge is not relevant, \eg~\emph{$<$front yards, has a, grass$>$} in \figref{fig:condition}, the introduced knowledge in this case actually creates a negative effect. As a result, the empirical errors will be further aggravated when optimizing with the answer prediction loss in \eqref{equ:lqik}. Tackling such a problem is non-trivial, as the `right knowledge' is made unknown to models. In view of this, we propose to address the above two issues by a novel supervised training scheme.

In the first step, we compute the loss with the image-question inputs and estimate the mathematical difference between this loss and the one with the retrieved knowledge, which is referred as the loss gap $\delta$,
\begin{equation}\label{equ:valid}
    \left\{ 
    \begin{aligned}
    \bar{\mathbf{y}} &= \text{MLP}(f^{be}(Q, I; \mathbf{\Theta})), \\
    \mathcal{L}_{QI} (Q, I; \mathbf{\Theta}) &= \sum_{j=1}^{\mid \Omega \mid} -y_j \log \bar{y}_j, \\
    \delta &=  \mathcal{L}_{QI} - \mathcal{L}_{QIK}.
    \end{aligned}
    \right.
\end{equation}

\textbf{Reader Loss.} Pertaining to \eqref{equ:valid}, if $\delta < 0$, it indicates that the knowledge becomes `noise' to the image-question input. In other words, the knowledge yields a negative influence for question answering, which however, should be ignored for optimization to avoid error propagation. In the light of this, we assign a regularization weight to the loss function in \eqref{equ:lqik} and have,
\begin{equation} \label{equ:reader}
    \mathcal{L}_{QIK} (Q, I, K_i; \mathbf{\Theta}) = \sigma (\delta) * \sum_{j=1}^{\mid \Omega \mid} -y_j \log \hat{y}_j,
\end{equation}
where $\sigma (\cdot)$ is the sigmoid function, which is a monotonically increasing function and utilized for smoothing the loss gap. In this way, the knowledge is more efficiently utilized by our model since only the knowledge with positive effect is taken into consideration. This scheme is expected to achieve a trade-off between accuracy and efficiency.

\subsection{Optimization}
Recall that for the Retriever, it is impractical to tell whether the retrieved knowledge is correct as there is no knowledge supervision for the benchmark dataset. Therefore, we compromise with the ground-truth answer, \ie~the exclusive signal in knowledge-based VQA, to perform pseudo knowledge labeling.

\textbf{Retriever Loss.} To achieve this goal, we reuse the loss gap $\delta$ as an indication for knowledge retrieval. In particular, we employ the tanh function to convert $\delta$ into the range of [-1, 1] to conform with the similarity interval. In this way, an optimal value 1.0 implies the retrieved knowledge matches perfectly with the given image-question pair, while a value -1.0 means the knowledge is totally not relevant. Accordingly, we design a Retriever loss $\mathcal{L}_{RET}$ to implement this idea, which is largely unexplored by previous literature,
\begin{equation} \label{equ:retriever}
    \mathcal{L}_{RET} (Q, I, \mathcal{K}; \mathbf{\Theta}) = \frac{1}{t}\sum_{i=1}^t\big(\text{tanh}(\delta) - \text{Sim}(QI, K_i)\big)^2,
\end{equation}
where $t$ denotes the number of retrieved knowledge entries and the similarity function can be referred to \eqref{equ:sim}. 

Finally, the overall loss is the linear combination of the Retriever loss $\mathcal{L}_{RET}$ and the Reader loss $\mathcal{L}_{QIK}$,
\begin{equation} \label{equ:final-loss}
    \mathcal{L} = \mathcal{L}_{QIK} + \lambda \mathcal{L}_{RET},
\end{equation}
where $\lambda$ serves as a hyper-parameter for balancing these two objectives. As can be observed, the whole framework can be trained in an end-to-end fashion.  

During inference, given the image-question pair, we first retrieve the most relevant knowledge entry by the Retriever. We then leverage this knowledge, together with the image and question, for answer prediction with the Reader.

\section{Experiments}\label{sec:experiments}

\subsection{Experimental Setting}
We elaborate the experimental protocols from the following four aspects: \emph{Dataset}, \emph{Evaluation Metric}, \emph{Multi-modal Implicit Knowledge Source} and the \emph{Implementation Details.}

\textbf{Dataset.} We conducted extensive experiments on the exclusive open-domain knowledge-based VQA benchmark -- OK-VQA~\cite{ok-vqa}. It consists of 14,055 questions and 14,031 images, where most questions are strictly encouraged to access external knowledge for the correct answer. In addition, the questions fall into eleven knowledge categories, such as \emph{vehicles and transportation (VT)} and \emph{brands, companies and products (BCP)}. 

\textbf{Evaluation Metric.} We adopted the standard metric in VQA and OK-VQA for evaluation~\cite{vqa,ok-vqa, focal}. For each predicted answer $a$, the accuracy is computed as,
\begin{equation*}\label{equ:metric}
  Accuracy = \text{min} \Big(1, \frac{\#\text{humans that provide answer $a$}}{3}\Big).
\end{equation*}
Note that each question is answered by ten annotators, and this metric takes the human disagreement into consideration~\cite{vqa, ok-vqa}.

\textbf{Multi-modal Implicit Knowledge Source.} We utilized three well-studied VLP backbones as the source of multi-modal implicit knowledge: VisualBert (VB)~\cite{visual-bert}, LXMERT~\cite{lxmert} and ViLT~\cite{vilt}, due to their diverse variety of settings. Among them, VisualBert and ViLT are single-stream methods, while LXMERT is a typical two-stream model. Besides, ViLT takes the image patch as visual inputs, while the others use the pre-trained regional features from UpDn~\cite{updn}.

\textbf{Implementation Details.} Compared with recent knowledge bases, ConceptNet~\cite{conceptnet} is large in its scale with ~8M nodes (larger than the recent CSKG~\cite{cskg}) and comprehensive with common sense knowledge, and we therefore leveraged it as a typical explicit knowledge source in this work. For this knowledge base, each triplet is flattened into a short sentence. After that, the pre-trained Roberta model~\cite{roberta} is employed to extract the sequential text features. Since VisualBert and LXMERT demand image features rather than image patch as visual inputs, we adopted the features extracted by~\cite{lxmert} for these two backbones. Pertaining to the hyper-parameters, we tuned the number of retrieved knowledge $t$ in \eqref{equ:retriever} from 1 to 5; the balancing weight $\lambda$ in \eqref{equ:final-loss} from 0 to 10 with a step size of 1. For all models, the learning rate is set to 1.0e-5 and decays $0.75\times$ with each training epoch. The AdamW optimizer~\cite{adamw} is employed for parameter optimization and the weight decay is fixed to 1.0e-4.
\vspace{-1em}
\subsection{Overall Results}

\begin{table}[!t]
    \centering
    \caption{Performance comparison on the FVQA dataset. The best performance is highlighted in bold.}
    \vspace{-1em}
    \begin{tabular}{l|c|c}
    \toprule
        Method                                      & Top-1 Acc (\%)     & Top-3 Acc (\%)     \\
    \midrule
    \midrule
        LSTM-Q+I~\cite{fvqa}                        & 22.97         & 36.76         \\
        \hspace{2mm} +Pre-VQA~\cite{fvqa}           & 24.98         & 40.40         \\
        Hie-Q+I~\cite{hie}                          & 33.70         & 50.00         \\
        \hspace{2mm} +Pre-VQA~\cite{hie}            & 43.14         & 59.44         \\
        BAN~\cite{ban}                              & 35.69         & -             \\
        \hspace{2mm} +KG-Aug~\cite{mm}              & 38.58         & -             \\
        Top1-QQmapping~\cite{fvqa}\hspace{5mm}      & 52.56         & 59.72         \\
    \midrule
        UnifER+LXMERT                               & 51.83         & 66.83         \\
        UnifER+ViLT                                 & \textbf{55.04}         & \textbf{69.72}         \\
    \bottomrule
    \end{tabular}
    \label{tab:fvqa}
\end{table}
\begin{figure}[!t]
  \centering
  \includegraphics[width=1.0\linewidth]{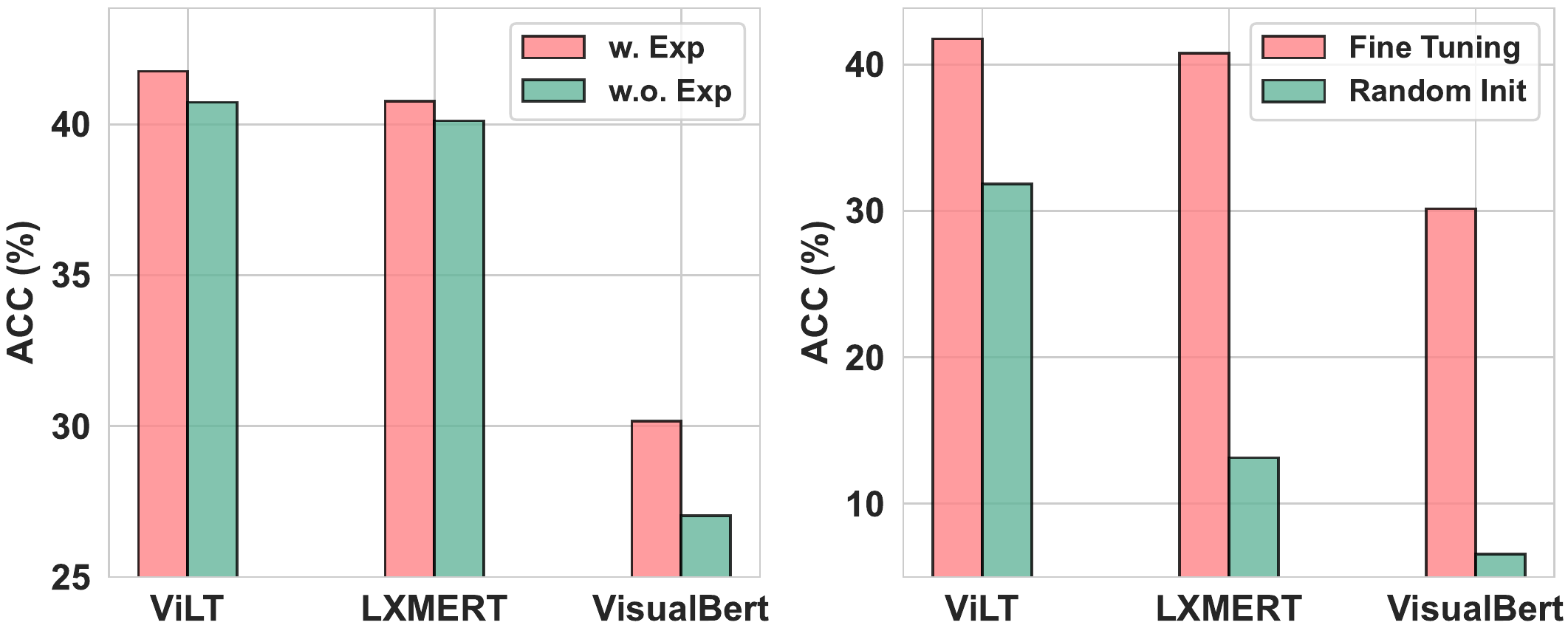}
  \vspace{-2em}
  \caption{Effectiveness validation of the explicit and implicit knowledge. Exp denotes explicit knowledge, and Fine Tuning represents the involvement of multi-modal implicit knowledge from pre-trained models.} \label{fig:bar}
  \vspace{-1em}
\end{figure}

The overall accuracy comparison of our method with knowledge-based VQA baselines is demonstrated in \tabref{tab:overall}. From this table, we have the following observations:
\begin{itemize}
    \item Conventional VQA approaches, \eg~BAN and MUTAN, perform less satisfactorily on this benchmark. It is because that these models are in essence knowledge free and thus not suited for  these knowledge-intensive questions. 
    \item More recent methods tend to introduce more knowledge to VQA models. Though these methods achieve certain advancement, there is a higher chance that they inject more noise, and the marginal performance gain can also be seen (MAVEx 39.20 v.s. KRISP 38.35).
    \item Our method with the ViLT backbone achieves the best over all the baselines. It is worth noting that unlike the recent methods, such as KRISP and MAVEx, our UnifER requires less explicit knowledge, \eg~without WP. The improvement can be attributed to the utilization of multi-modal implicit knowledge as well as the noise reduction scheme presented by our method.
\end{itemize}
In addition, we also conducted experiments on  the FVQA~\cite{fvqa} dataset, which provides ground-truth knowledge entry for each image-question pair, for completeness. As can be observed in \tabref{tab:fvqa}, our method can surpass previous models by a large margin, especially on the Top-3 Acc metric.

\begin{table}[!t]
    \centering
    \caption{Effectiveness of smoothing function for both the Retriever relevance labeling and Reader instance selection.} 
    \vspace{-1em}
    \resizebox{0.95\columnwidth}{!}{%
    \begin{tabular}{l|cc|cc|c}
    \toprule
        \multirow{2}{*}{Backbone}       & \multicolumn{2}{c|}{Function} & \multicolumn{2}{c|}{Scope} & Accuracy  \\
        \cline{2-5}
                                        & Smoothing     & Binary        & Retriever     & Reader        & (\%)            \\
    \midrule
    \midrule
        \multirow{3}{*}{VisualBert}     &               & \checkmark    &               & \checkmark    & 17.89     \\
                                        &               & \checkmark    & \checkmark    &               & 26.75     \\
                                        & \checkmark    &               & \checkmark    & \checkmark    & 30.16     \\    
    \midrule
        \multirow{3}{*}{LXMERT}         &               & \checkmark    &               & \checkmark    & 26.87     \\
                                        &               & \checkmark    & \checkmark    &               & 34.67     \\
                                        & \checkmark    &               & \checkmark    & \checkmark    & 40.94     \\ 
    \midrule
        \multirow{3}{*}{ViLT}           &               & \checkmark    &               & \checkmark    & 32.95     \\
                                        &               & \checkmark    & \checkmark    &               & 37.42     \\
                                        & \checkmark    &               & \checkmark    & \checkmark    & 42.13     \\ 
    \bottomrule
    \end{tabular}}
    \label{tab:binary}
    \vspace{-1em}
\end{table}

\begin{figure}[!t]
  \centering
  \includegraphics[width=1.0\linewidth]{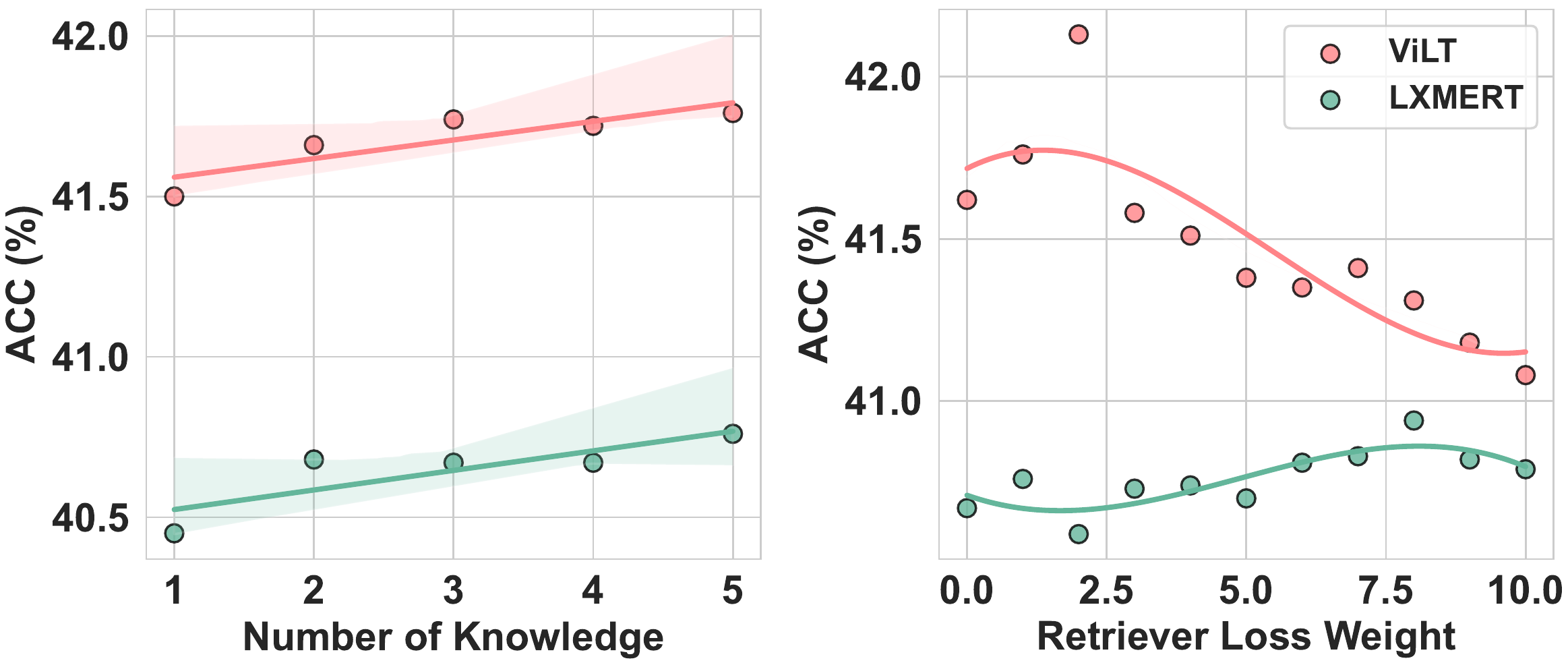}
  \vspace{-2em}
  \caption{Accuracy variation with respect to the number of retrieved knowledge and the retriever loss weight.} \label{fig:number}
  \vspace{-1em}
\end{figure}

\begin{figure}[!t]
  \centering
  \includegraphics[width=1.0\linewidth]{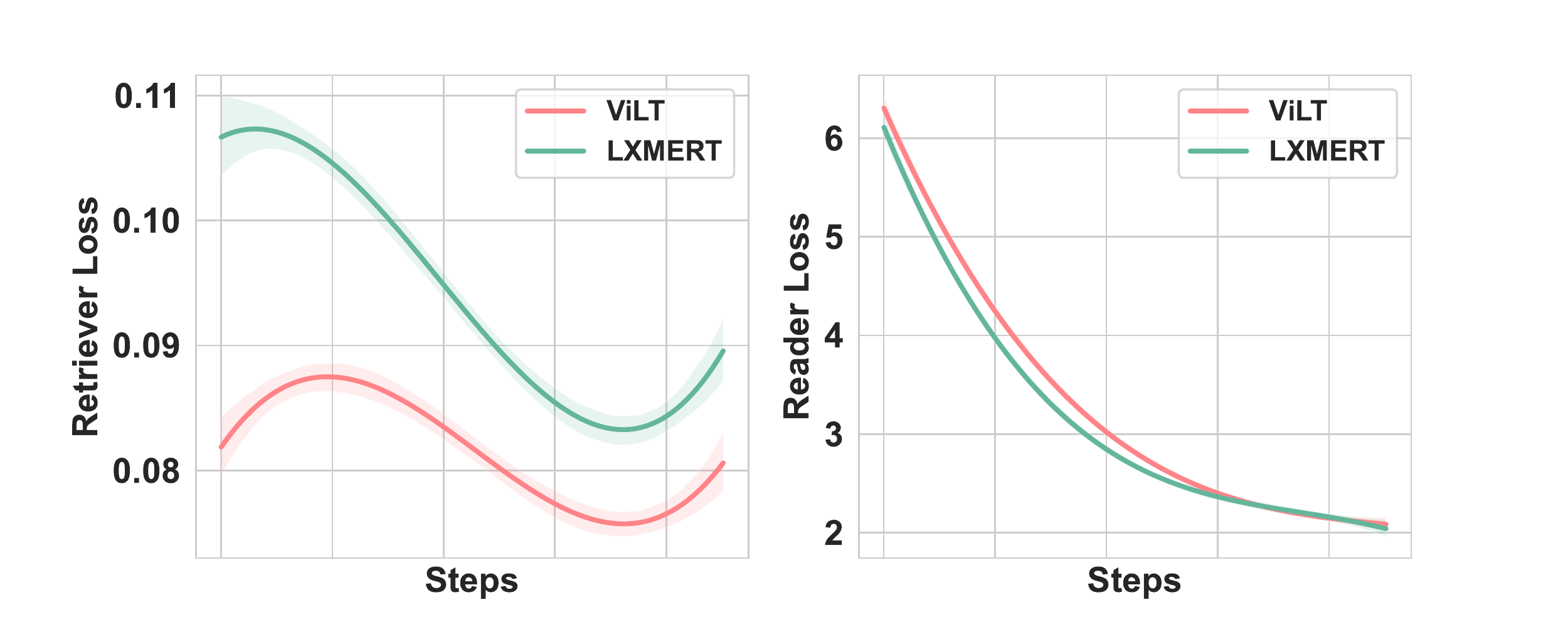}
  \vspace{-2em}
  \caption{Convergence analysis of Retriever and Reader Losses with respect to training steps.} \label{fig:converge}
   \vspace{-2em}
\end{figure}

\subsection{Ablation Study} 
We studied the effectiveness of various components from our method with the following research questions.
All experiment are conducted with OK-VQA~\cite{ok-vqa}.

\textbf{RQ1}: Do both explicit and implicit knowledge contribute to the final answer prediction?

To answer this question, we first eliminated the explicit knowledge, \ie~ConceptNet knowledge base, to observe the performance change. As demonstrated in \figref{fig:bar}, the model degrades to some extent when removing the explicit knowledge from our method (w.o. Exp), especially for the VisualBert backbone. By contrast, after we masked the implicit knowledge, namely, randomly initializing all the model parameters (Random Init), we can observe a drastic performance drop on all these three backbones. This experiment reveals that the implicit knowledge plays a more important role than its explicit counterpart, and one possible reason is that the models pre-trained on other image-text datasets can provide certain knowledge for downstream question answering objective. 


\textbf{RQ2}: How do the hyper-parameters, \ie~$n$ in \eqref{equ:retriever} and $\lambda$ in \eqref{equ:final-loss}, and the smoothing setting affect the model performance?

It is reasonable that retrieving more knowledge for Reader can improve the answering confidence, yet introducing higher training overhead and memory cost. \figref{fig:number} shows that for both LXMERT and ViLT backbones, 
increasing the number of retrieved knowledge monotonously improve the model performance.
Next, we explored the influence of $\lambda$, which balances the learning of Retriever and Reader. As observed, ViLT prefers a smaller weight for Retriever, \ie~2, while LXMERT performs better on larger weights.  

Note that we employed a smoothing setting for both the Retriever relevance labeling and Reader instance selection in \eqref{equ:retriever} and \eqref{equ:reader}, respectively. It is still questionable whether the binary deterministic setting works in these two cases. To investigate this, we replaced the tanh function in \eqref{equ:retriever} with the $\{-1, 1\}$ value by a threshold to observe the model performance change. Similar setting is also employed on \eqref{equ:reader} to replace $\sigma$ with $\{0, 1\}$. As can be seen in \tabref{tab:binary}, the performance deteriorates drastically when changed to the the binary settings. Besides, the Reader instance selection is more influenced than that of the Retriever relevance labeling. It is mainly because that many inapplicable knowledge entries make the training instance invalid, thereby shrinking the training samples and leading to the potential under-fitting problem.

\begin{figure}[!t]
  \centering
  \includegraphics[width=1.0\linewidth]{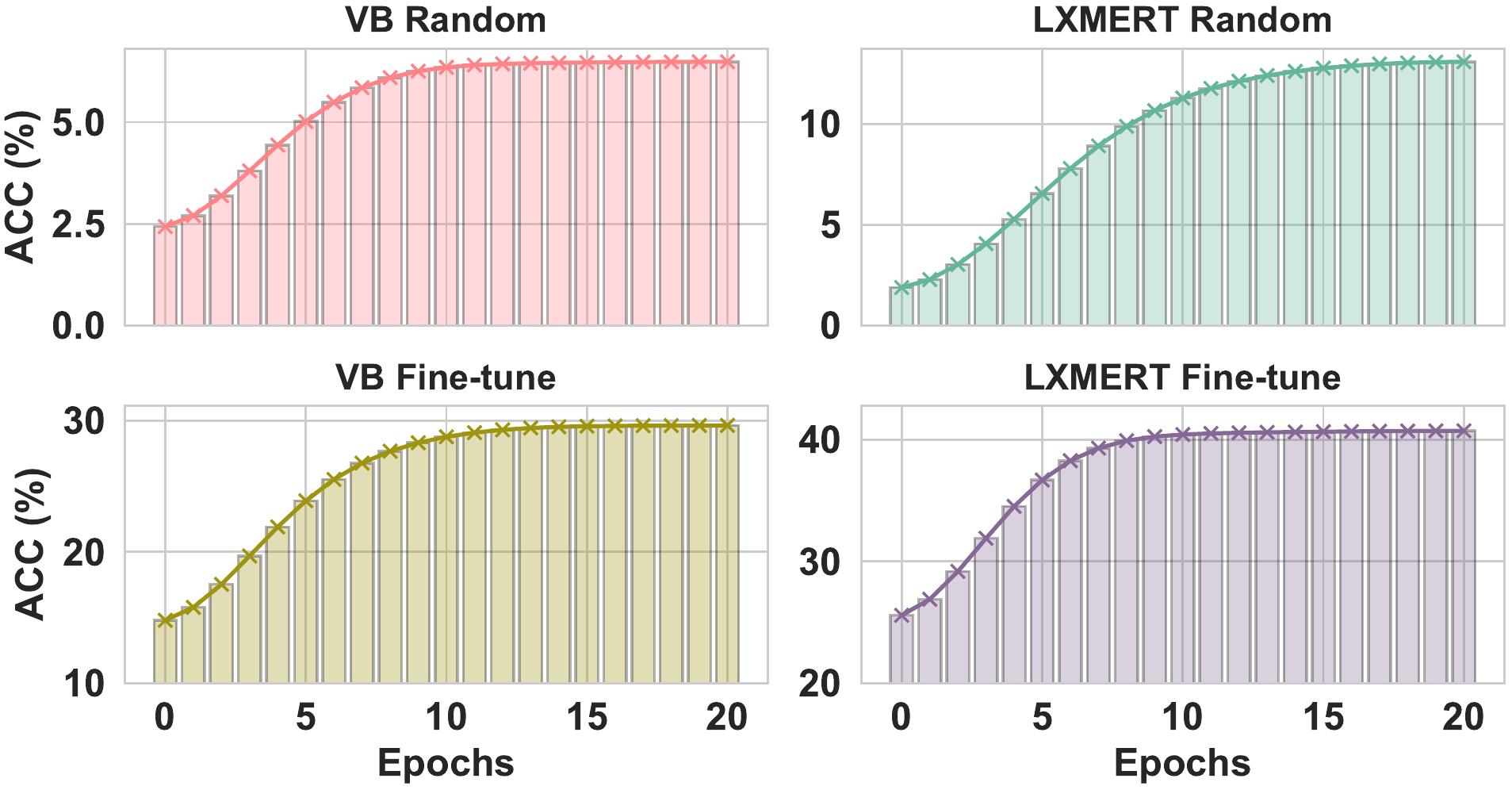}
  \vspace{-2em}
  \caption{Convergence analysis of accuracy (\%) for fine-tuned and randomly initialized models with respect to epochs.} \label{fig:mix}
  \vspace{-2em}
\end{figure}

\textbf{RQ3}: What makes the dominant factors in model training convergence?

\begin{figure*}[!t]
  \centering
  \includegraphics[width=1.0\linewidth]{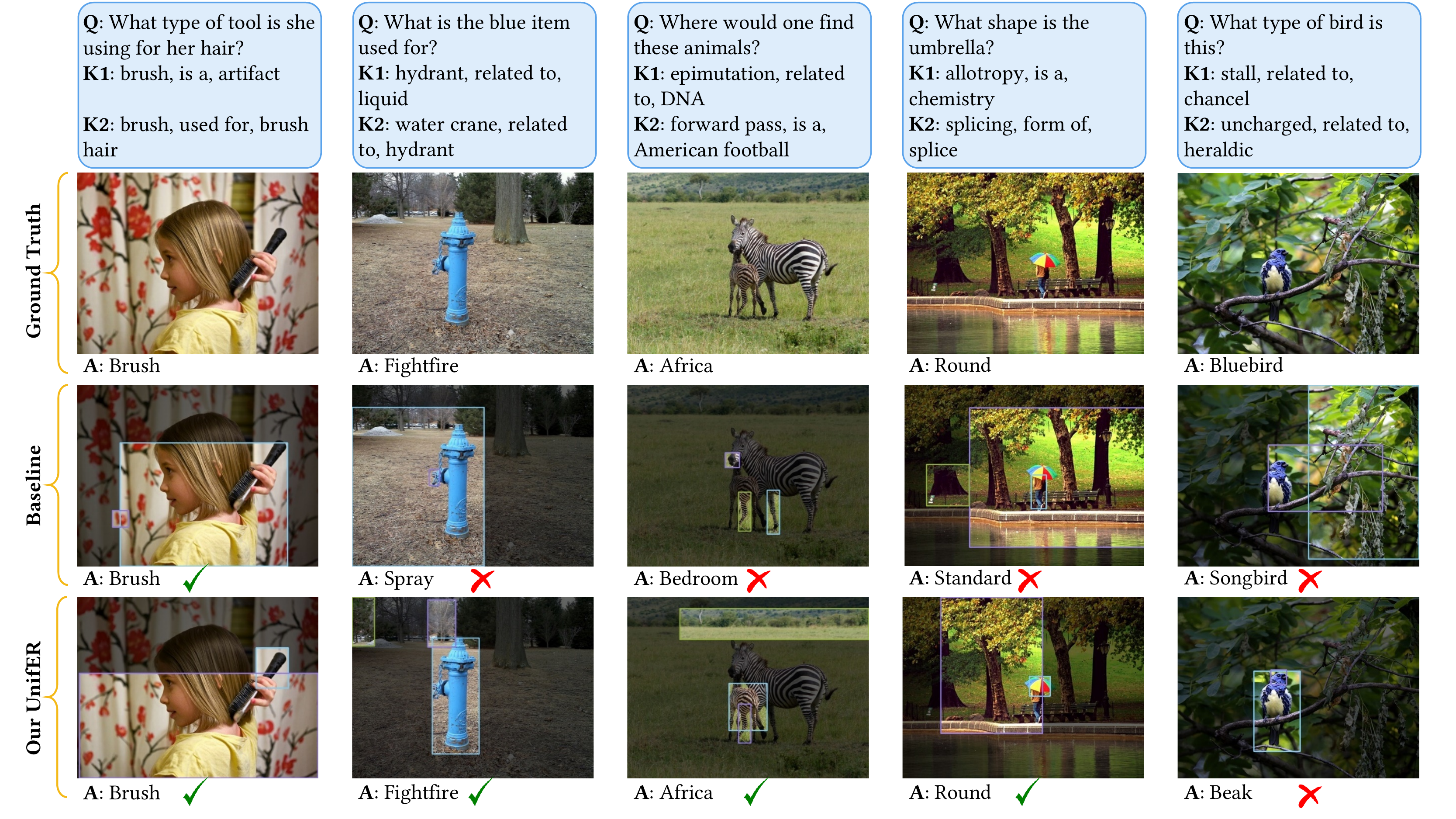}
  \vspace{-2em}
  \caption{Qualitative results on five cases. From top to bottom, we illustrate the given question, retrieved two entries of knowledge, ground-truth answer with the original image, attention map with the predicted answer from the baseline~\cite{lxmert}, and attention map with the predicted answer from our method.} \label{fig:viz}
  \vspace{-1em}
\end{figure*}

We aimed to address this question with two sets of factors. The key findings for this experiment are shown in \figref{fig:converge} and \figref{fig:mix}. The first factor is the Retriever model supervision. As discussed in previous sections, current models put no supervision for knowledge retrieval since there is no such relevant knowledge ground-truth. As a result, the retrieved knowledge can be inapplicable for answering, leading to error-propagation problem and sub-optimal model performance. We observe in \figref{fig:converge} that the Reader loss of ViLT and LXMERT is similar with more training steps. However, there exists a gap between their Retriever losses, which results in their overall accuracy difference, \ie~42.13 v.s. 40.94. This finding proves that a better Retriever model does benefit the answer prediction in Reader. In addition, we attributed the second factor to the incorporation of multi-modal implicit knowledge. \figref{fig:mix} demonstrates that when using the pre-trained model, \ie~Fine-tune, the model performance exhibits a faster convergence than the one without.

\subsection{Qualitative Analysis}
We performed detailed case study with the LXMERT backbone~\cite{lxmert} and visualized some results in \figref{fig:viz}. Among these five cases, the first two are with positive retrieved knowledge entries. For instance, the knowledge \emph{$<$brush, used for, brush hair$>$} is useful for obtaining the right answer \emph{brush}; \emph{$<$hydrant, related to, liquid$>$} helps the second question to predict the \emph{fightfire} with a high confidence. In addition, the attention maps from our method are more human-friendly regarding these two cases. Pertaining to the middle two cases, the retrieved explicit knowledge is only partially or totally not relevant with the given questions. Thanks to the multi-modal implicit knowledge, our method can predict the right answers for these two. The possible reasons are two-fold: 1) the backbone model has learned the implicit  that \emph{zebra is from Africa} for the third case and \emph{umbrella is round} for the fourth case; and 2) our method puts more attention on the right image region as opposed to the baseline (\eg~\emph{umbrella} for the forth case). The last one is a failure case from our method. In fact, this question might beyond the capability of the backbone model as it has not seen such fine-grained classes from the existing multi-modal datasets. It thus brings our attention to design more powerful mechanisms to retrieve the desired explicit knowledge. 
\section{Conclusion and Future Work}\label{sec:conclusion}
In this paper, we propose a unified end-to-end retriever-reader framework towards knowledge-based VQA. Our work stands out from the existing studies by two merits. First, we well study the superiority of  multi-modal implicit knowledge from pre-trained vision-language models. Second, a novel scheme is carefully designed to address the error propagation problem originated from the explicit knowledge utilization. We perform extensive experiments on the benchmark dataset and demonstrate that our method achieves significant performance improvement over baselines. 

Our empirical findings suggest two potential future directions for knowledge-based VQA: 1) building more capable and robust vision-language pre-training models and transferring them to the downstream knoweledge-based VQA, and 2) developing more beneficial formulations for structured explicit knowledge to make it more applicable. In addition, telling models what questions cannot be addressed with image content only contributes another interesting research topic.  

\begin{acks}
This research is supported by the National Research Foundation, Singapore under its Strategic Capability Research Centres Funding Initiative. Any opinions, findings and conclusions or recommendations expressed in this material are those of the author(s) and do not reflect the views of National Research Foundation, Singapore.
\end{acks}

\bibliographystyle{format/ACM-Reference-Format}
\bibliography{UnifER}

\end{document}